%% file: main.tex
\pdfoutput=1

\documentclass[11pt]{article}

\usepackage[]{acl}

\usepackage{times}
\usepackage{latexsym}

\usepackage[T1]{fontenc}

\usepackage[utf8]{inputenc}

\usepackage{microtype}

\usepackage{inconsolata}

\usepackage{soul}
\usepackage[textsize=small, disable]{todonotes}
\usepackage{xspace}
\usepackage{multirow}
\usepackage{booktabs}

\usepackage{amsmath}
\usepackage{algpseudocode}

\usepackage{tabularray}
\usepackage{amsfonts}
\usepackage{pifont}
\usepackage{arydshln}
\usepackage{algorithm}
\usepackage{algorithmicx}
\usepackage{hyperref}
\usepackage{graphicx}
\usepackage{subcaption}

\usepackage{mathtools}

\DeclarePairedDelimiter\floor{\lfloor}{\rfloor}
\newcommand{\proj}{Shears\xspace}

\usepackage{tikz}

\title{
\proj: Unstructured Sparsity with Neural Low-rank Adapter Search 
}

\author{J. Pablo Muñoz\thanks{Co-first authors.} \\
  Intel Labs \\
  \texttt{pablo.munoz@intel.com} \\\And
  Jinjie Yuan\footnotemark[1] \\
  Intel Corporation \\
  \texttt{jinjie.yuan@intel.com} \\\And
  Nilesh Jain \\
  Intel Labs \\
  \texttt{nilesh.jain@intel.com} \\}

\begin{document}
\maketitle

\input{content/0_abstract}

\input{content/1_intro}

\input{content/2_preliminaries}

\input{content/3_methodology}

\input{content/4_experiments}

\input{content/5_conclusion}

\bibliography{content/main} 
\clearpage
\appendix
\input{content/appendix}

\end{document}

%% file: content/0_abstract.tex
\section*{Abstract}

Recently,
several approaches 
successfully demonstrated that weight-sharing Neural Architecture Search (NAS) can effectively explore a search space of elastic low-rank adapters (LoRA), allowing the parameter-efficient fine-tuning (PEFT) and compression of large language models. 
In this paper, we introduce a novel approach called \textbf{\proj}, demonstrating how the integration of cost-effective sparsity and a proposed Neural Low-rank adapter Search (NLS) algorithm can further improve the efficiency of PEFT approaches. 
Results demonstrate the benefits of \proj compared to other methods, reaching high sparsity levels while improving or with little drop in accuracy, utilizing a single GPU for a pair of hours. 

%% file: content/1_intro.tex
\section{Introduction}
Large language models (LLMs) exhibit impressive capabilities in comprehensive language understanding, as evidenced by their remarkable zero-shot generation across various tasks.
However, supervised fine-tuning is often employed to unlock their true potential in real-world applications.
Fine-tuning 
is essential for tailoring performance to domain-specific or proprietary data, bridging the gap between general language understanding and task-specific precision.
Recently, parameter-efficient fine-tuning (PEFT) \cite{ding2022delta} has emerged as a crucial strategy for efficiently boosting the performance of LLMs in domain-specific tasks.

In addition to fine-tuning, 
increasing 
model parameters is another critical strategy to improve model performance.
LLMs have produced impressive achievements as they scale to significant sizes, such as PaLM with 540 billion parameters \cite{chowdhery2022palm}. 
The 
projection for 
future models suggests a continuous escalation in parameter count, anticipating improved performance.
However, this trend also underscores the growing demands on computing devices. As model parameters increase, so does the computational complexity, necessitating more powerful hardware and infrastructure. 
In this context, the importance of model compression becomes particularly evident.
Model compression 
is 
a crucial strategy to mitigate 
these 
challenges and make LLMs more accessible and deployable across a broader spectrum of devices.

Motivated by the 
significance of PEFT and model compression, this paper introduces a novel approach called \textbf{\proj}, showing the effective integration of PEFT and model compression to optimize the LLM performance  
with a high 
sparsity level. 
In the proposed methodology, we initiate the process by employing a zeroth-order sparse approach to induce sparsity 
in the LLM.
Subsequently, we introduce elastic low-rank adapters into the sparsified model and apply Neural Low-rank adapter Search (NLS) to train a super-adapter network. 
Finally, a search algorithm is employed to identify the optimal adapter configuration.
The contributions of this work can be summarized as follows:
\begin{enumerate}
    \item We propose a practical solution 
    combining 
    model compression and PEFT, manifested in cost-effective sparsity and the proposed neural low-rank adapter search.
    \item Our approach
    features 
    three well-designed steps, i.e., unstructured sparsification, super-adapter training, and sub-adapter search. The proposed approach effectively obtains sparse fine-tuned LLMs that reduce inference time. 
    \item Experiments and ablation studies to confirm that our approach can produce models that maintain high accuracy while significantly increasing their sparsity levels. 
\end{enumerate}

The content of this paper uses the following outline: Section \ref{sec:preliminaries} discusses the algorithms \proj uses. Section \ref{sec:method} describes the three stages and details of our practical solution. Section \ref{sec:experiments} presents results with several models on a variety of tasks. We offer concluding remarks in Section \ref{sec:conclusion}, and due to space limitations, we provide more details and a \emph{Related Work} section in the Appendix.

%% file: content/2_preliminaries.tex
\begin{figure*}[th]
\includegraphics[width=0.8\textwidth]{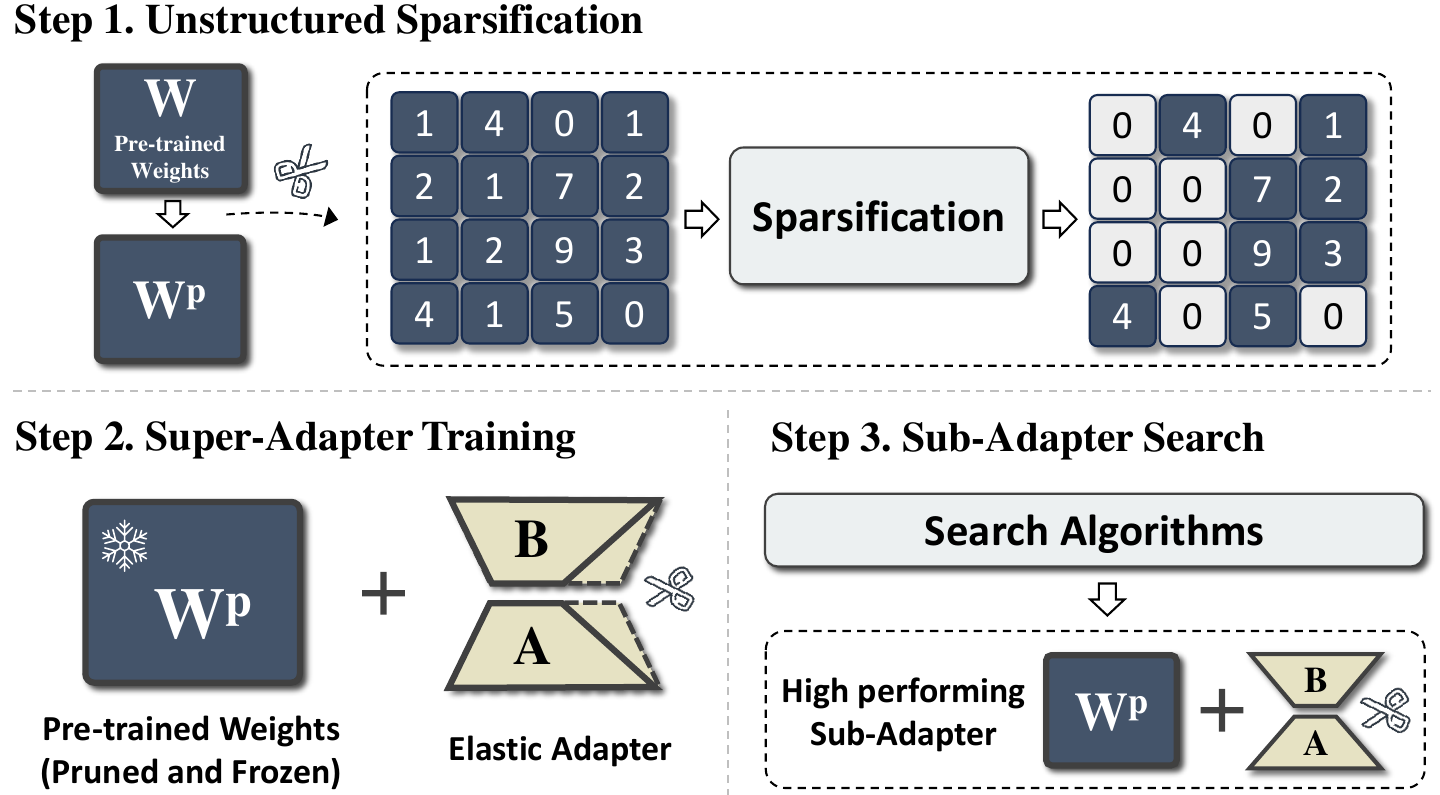}
\centering
\caption{\proj workflow. Initially, \proj employs a zeroth-order pruning algorithm to induce sparsity in the given LLM. Subsequently, the framework generates a super-adapter network trained by activating subnetworks within the search space of elastic adapters. Finally, \proj yields sub-adapter networks that exhibit high performance.
}
\label{fig:workflow}
\end{figure*}

\section{Preliminaries}
\label{sec:preliminaries}

\subsection{Sparsification} 
\label{sec:sparsification}

Our approach introduces 
sparsity into LLMs using a zeroth-order and cost-effective 
algorithm.
In our experiments, we utilized the Wanda algorithm \cite{sun2023wanda}, which calculates weight importance based on 
weights, and activations and then leverages this information for unstructured pruning. 
Specifically, given a weight matrix $\boldsymbol W$ and input feature activations $\boldsymbol X$, Wanda computes the weight importance $\boldsymbol S$ as the element-wise product of the weight magnitude and the norm of input activations, formulated as follows:
\begin{equation}
\boldsymbol S = |\boldsymbol W| \cdot \|\boldsymbol X\|_2. 
\label{eq:wanda}
\end{equation}
Wanda compares the weight importance scores within each row in $\boldsymbol W$.
After obtaining the importance information, 
the algorithm zeroes out 
the less 
critical 
weights according to the specified sparsity level.
The sparsification approach efficiently obtains a model with any level of unstructured sparsity desired before training.

\subsection{Low-Rank Adaptation}
Recently, PEFT technology has emerged as a solution to address 
the challenges of fine-tuning large-scale models.
Among PEFT approaches, Low-Rank Adaptation (LoRA) \cite{hu2022lora} has shown notable efficacy in fine-tuning Transformer-based models for downstream NLP tasks. LoRA constraints the update for a pre-trained weight, $\boldsymbol W_0\in \mathbb{R}^{d\times k}$, 
by a low-rank decomposition $\boldsymbol W_0+ \Delta \boldsymbol W = \boldsymbol W_0+ \boldsymbol B\boldsymbol A$, where $\boldsymbol B\in \mathbb{R}^{d\times r}, \boldsymbol A\in \mathbb{R}^{r\times k}$, and the rank $r \ll \min(d,k)$.
Throughout the training process, $\boldsymbol W_0$ remains frozen and does not undergo gradient updates, while only the parameters of $\boldsymbol A$ and $\boldsymbol B$ are trained.
For the linear projection, $\boldsymbol H = \boldsymbol W_0 \boldsymbol X$, the forward pass with LoRA is formulated as follows:
\begin{equation}
\boldsymbol H = \boldsymbol W_0 \boldsymbol X + \Delta \boldsymbol W \boldsymbol X = \boldsymbol W_0 \boldsymbol X + \boldsymbol B\boldsymbol A \boldsymbol X,
\label{eq:lora}
\end{equation}
where $\boldsymbol A$ is initialized with a random Gaussian while $\boldsymbol B$ is initialized with zeros, ensuring $\Delta \boldsymbol W =\boldsymbol B\boldsymbol A$ is zero at the beginning of training.
Inspired by this approach, this paper integrates elastic LoRA adapters into Neural Architecture Search.

%% file: content/3_methodology.tex
\section{Methodology}
\label{sec:method}
In this section, we delve into the proposed approach, \proj.
Figure \ref{fig:workflow} illustrates the overview of the \proj pipeline.
As depicted in the figure, the method comprises three key steps: i) Unstructured Sparsification, ii) Super-Adapter Training, and iii) Sub-Adapter Search.
Through these steps, the model undergoes sparsification and neural low-rank adapter search while preserving a performance comparable to the fine-tuned model from the original model. Next, we discuss the details of each step.

\subsection{Unstructured Sparsification}

As illustrated in step 1 of Figure \ref{fig:workflow}, 
\proj employs a sparsification metric to zero out
the less essential weights of
the given LLM. 
As mentioned in Section \ref{sec:sparsification}, we apply the Wanda algorithm (Equation \ref{eq:wanda}) in our main experiments.
However, in theory, \proj could utilize other algorithms, e.g., movement sparsity \cite{sanh2020movement} or SparseGPT \cite{frantar-sparsegpt}. 
The pruned weights $\boldsymbol{W_p}$ are kept frozen throughout the subsequent stages of the overall pipeline. 
In this step, we factor in the cost of obtaining the weight importance structure. When using Wanda, only a tiny subset of inputs needs to forward pass to get the unstructured importance measurements instead of more sophisticated approaches that require weight updates and training iterations. The reader can find further details about the Wanda algorithm in its paper \cite{sun2023wanda}. Obtaining $\boldsymbol{W_p}$ for a model with seven billion parameters takes less than five minutes on a single GPU, as utilized in our experiments. 

\subsection{
Super-Adapter Training
}

Subsequently, a weight-sharing super-adapter network is generated using the space of low-rank adapters. 
\proj does not make the original model weights $\boldsymbol{W}$ elastic as opposed to the elastic configurations of the adapters. 
The super-adapter network is then fine-tuned for a particular task through Neural Low-Rank Adapter Search (NLS), which we discuss next.

\paragraph{Neural Low-Rank Adapter Search (NLS)} An elastic low-rank adapter can have numerous possible configurations. NLS leverages the mechanisms inherited from Neural Architecture Search (NAS) to activate adapter configurations and proceed with the forward and backward passes to fine-tune the possible sub-adapters. After fine-tuning the super-adapter network, which takes a pair of hours in a single GPU (further details in Section \ref{sec:experiments}), \proj discovers a configuration that yields comparable accuracy on the target task. 

\subsection{
Sub-Adapter Search
}

Identifying an optimal sub-adapter configuration can be an expensive endeavor. Although the search 
space of elastic adapter configurations is significantly smaller than if we also include subnetworks derived from the pre-trained weights of the LLM, the number of possible configurations for the adapters is still considerable. Sampling and evaluating these configurations can demand a significant amount of time. 
We can employ several approaches to explore search spaces of neural network configurations, such as evolutionary search using the Non-Dominated Sorting Genetic Algorithm II (NSGA-II) \cite{nsgaII} or a variation like RNSGA-II \cite{rnsga-ii}. 
However, the cost of this type of search in LLM is prohibitive. To address this, we employ two alternatives. First, we extract a sub-adapter configuration using a heuristic. Then, suppose the performance of this configuration falls short of the desired outcome, 
a well-designed hill-climbing algorithm can be utilized to search for better configurations. Concretely,
\proj can initiate a hill-climbing algorithm from the sub-adapter configuration found with the heuristic to explore its neighborhood and discover potentially improved configurations. This search approach is significantly less expensive than other search strategies, e.g., evolutionary search. 
Formally, the heuristic strategy, initially proposed in LoNAS \cite{lonas2024}, 
to obtain a reference subnetwork configuration approximately at the center of the search space is as follows:
\begin{equation}
\small
    \textbf{\proj-Heuristic}_{l_i} \leftarrow \textbf{\proj-Maximal}_{l_i}[c],  \text{s.t. } 
    c = \floor*{\frac{n}{2}},
\label{eq:heuristic}
\end{equation}
where $c$ represents the index of the elastic width (rank of the adapter) configuration for the adapter $l_i$, chosen from a total of $n$ possible elastic configurations at that adapter. This heuristic provides a (weak) indication of the performance of smaller sub-adapter networks. 

\begin{table*}[htb]
\setlength{\belowcaptionskip}{5pt}
\setlength{\tabcolsep}{4.5pt}
\caption{Sparsity and test accuracy (\%) comparison of \proj with other LLM-Adapter approaches. The baseline results are those reported by \citet{hu2023llm_adapters}. \proj models have high accuracy while significantly increasing model sparsity.}
\footnotesize
\centering
\renewcommand\arraystretch{1.2}
\begin{tabular}{llcccccccc}
\toprule
\multirow{2}{*}{\textbf{LLM}} & \multirow{2}{*}{\textbf{Method}} & \multirow{2}{*}{\textbf{Sparsity}}  & \multicolumn{4}{c}{\textbf{Datasets | Accuracy(\%)}} & \multirow{2}{*}{\textbf{Average}} \\ 
& &   & \textbf{GSM8K} & \textbf{AQuA} & \textbf{MAWPS} & \textbf{SVAMP} & & \\
\midrule
GPT-3.5 &  Zero-shot CoT
& -  & {56.4}& {38.9} & {87.4} & {69.9} & 70.4\\
\midrule
\multirow{6}{*}{LLaMA$_{\text{7B}}$}
& Prefix & -  & 24.4 & 14.2 & 63.4 & 38.1 & 35.0 \\
& Series & -  & 33.3 & 15.0 & 77.7 & \textbf{52.3} & 44.6 \\
& Parallel & -  & 35.3 & 18.1 & 82.4 & 49.6 & 46.4 \\
& LoRA & - & \textbf{37.5} & 18.9 & 79.0 & 52.1 & \textbf{46.9} \\			

\cdashline{2-9}
& \textbf{\proj} & \textbf{40\%} & 36.8 & 19.7 & \textbf{83.2} & 47.7 & \textbf{46.9}  \\ 
& \textbf{\proj} & \textbf{50\%} & 36.1 & \textbf{22.0} & 78.6 & 44.5 & 45.3 \\ 
\midrule
\multirow{6}{*}{LLaMA$_{\text{13B}}$}
& Prefix & - &	31.1&		15.7&	66.8&	41.4&	38.8  \\
& Series & - &	44.0&		\textbf{22.0}&	78.6&	50.8&	48.9 \\
& Parallel & -  &	43.3&	20.5&	81.1&	\textbf{55.7}&	50.2\\
& LoRA & - &	47.5&		18.5&	\textbf{83.6}&	54.6&	51.1 \\
\cdashline{2-9}
& \textbf{\proj} & \textbf{40\%} & \textbf{48.3} & 21.3 & 83.2 & 55.2 & \textbf{52.0}  \\ 
& \textbf{\proj} & \textbf{50\%} & 45.1 & \textbf{22.0} & 83.2 & 53.3 & 50.9 \\ 

\bottomrule
\end{tabular}
\label{tab:main_results_llama}
\end{table*}

%% file: content/4_experiments.tex
\begin{table*}[hbt]
\setlength{\belowcaptionskip}{5pt}
\setlength{\tabcolsep}{2.0pt}
\caption{Sparsity and test accuracy (\%) comparison of \proj with other LLM-Adapter approaches on commonsense reasoning datasets.
The result of {zero-shot}$^{1}$ is derived from \citet{touvron2023llama}, and the result of {zero-shot}$^{2}$ is from LLM-Pruner \cite{ma2023llmpruner}. LLM-Pruner employs prompts different from those used by \citet{touvron2023llama} for zero-shot evaluation since they do not provide the prompts they used.
Almost all results of the PEFT baselines are obtained from \citet{hu2023llm_adapters}, except for the LoRA baseline in the 15k train dataset, which we experimented with the official implementation.
}
\footnotesize
\centering
\renewcommand\arraystretch{1.4}
\begin{tabular}{lclccccccccccccc}
\toprule
\multirow{2}{*}{\textbf{LLM}} & \textbf{Train} & \multirow{2}{*}{\textbf{Method}} & \multirow{2}{*}{\textbf{Sparsity}}  & \multicolumn{8}{c}{\textbf{Datasets | Accuracy(\%)}} & \multirow{2}{*}{\textbf{Average}} \\ 
& \textbf{Set Size} &  &   & \textbf{BoolQ} & \textbf{PIQA} & \textbf{SIQA} & \textbf{HellaSwag} & \textbf{WinoG}& \textbf{ARC-e}& \textbf{ARC-c}& \textbf{OBQA}& & \\
\midrule
GPT-3.5 & - & - & - & 73.1 &	85.4	&68.5&	78.5&	66.1	&89.8	&79.9&	74.8&	77.0 \\
\midrule

\multirow{11}{*}{LLaMA$_{\text{7B}}$}
& - & {Zero-shot}$^{1}$ & - & 76.5& 79.8& 48.9 &76.1 &70.1 &72.8 &47.6 & 57.2 & 66.1 \\
& - & Zero-shot$^{2}$ & - & 73.2 &78.4& 32.9  & 73.0& 67.0 &67.5 &41.4 &42.4 & 59.5\\
\cline{2-13}
& \multirow{3}{*}{15k} & LoRA\raisebox{0.5ex}{*} & - & 62.6&	75.3&	67.9&	52.9&	58.6&	\textbf{79.2}&58.3&	71.2&	65.8 \\	
\cdashline{3-13}
& & \textbf{\proj} & \textbf{40\%} & \textbf{65.5}	&\textbf{76.0}&	\textbf{71.2}&	\textbf{56.8}	&\textbf{65.6}&	79.0&	\textbf{62.6}&	\textbf{76.4}	&\textbf{69.1} \\ 
& & \textbf{\proj} & \textbf{50\%} & 62.5	&75.7	&69.7&	54.8&	65.7&	75.1	&59.5&	72.6&	66.9 \\  
\cline{2-13}
& \multirow{6}{*}{170k} & Prefix & -  & 64.3&	76.8	&73.9&	42.1&	72.1&	72.9	&54.0&	60.6&	64.6 \\
& & Series & -  & 63.0&	79.2&	76.3&	67.9&	75.7	&74.5	&57.1&	72.4&	70.8 \\
& & Parallel & -  & 67.9&	76.4	&\textbf{78.8}&	69.8&	\textbf{78.9}&	73.7&	57.3	&75.2&	72.3 \\
& & LoRA & - & \textbf{68.9}	&\textbf{80.7}	&77.4&	78.1&	78.8&	\textbf{77.8}&	61.3&	74.8&	74.7 \\			
\cdashline{3-13}
& & \textbf{\proj} & \textbf{40\%} & 67.0	&79.9	&76.6&	\textbf{80.1}&	78.6&	76.9&	\textbf{62.3}&	\textbf{77.8}	&\textbf{74.9} \\ 
& & \textbf{\proj} & \textbf{50\%} & 67.3&	79.1&	77.5	&73.3&	77.7	&74.4&	57.9&	72.8&	72.5 \\  
\bottomrule
\end{tabular}
\label{tab:commonsense170k_results}
\end{table*}
\section{Experiments}
\label{sec:experiments}
\proj is implemented by extending BootstrapNAS \cite{bootstrapnas_aaai} and OpenVINO's Neural Network Compression Framework\footnote{https://github.com/openvinotoolkit/nncf}. 
We explore the benefits of \proj by generating and fine-tuning super-adapter networks for various LLMs. 
The following sections detail our experimental setup and the analysis of the results.

\subsection{Experimental Setup}

\paragraph{Datasets.} 
Following the work of LLM-Adapters \cite{hu2023llm_adapters} \footnote{https://github.com/AGI-Edgerunners/LLM-Adapters}, 
we assess the performance of \proj across a diverse range of tasks, including four math reasoning datasets (GSM8K \cite{cobbe2021training_GSM8K}, AQUA \cite{ling-etal-2017-program-aqua}, MAWPS \cite{lan2022mwptoolkit} and SVAMP \cite{patel-etal-2021-nlp-svamp}) and eight commonsense reasoning datasets (BoolQ \cite{clark2019boolq}, PIQA \cite{Bisk2020_piqa}, SIQA \cite{sap-etal-2019-social_siqa}, HellaSwag \cite{zellers2019hellaswag}, WinoGrande \cite{winogrande}, ARC \cite{Clark2018ThinkYH_arc} and OBQA \cite{Mihaylov2018CanAS_obqa}).
Leveraging GPT-3.5, the LLM-Adapters team generated high-quality, unified datasets for training while compiling several math or commonsense datasets.
Additionally, we conduct evaluations of \proj on the original GSM8K training dataset for comparison with the work of \citet{kurtic2023sparse}.

\paragraph{Models.}
We validate our approach using the LLaMA-series \cite{touvron2023llama} and MPT-series \cite{MosaicMPT7b} language models.
Specifically, we generate \proj super-adapter networks from LLaMA$_{\text{7B}}$\footnote{https://huggingface.co/yahma/llama-7b-hf}, LLaMA$_{\text{13B}}$\footnote{https://huggingface.co/yahma/llama-13b-hf}, and MPT$_{\text{7B}}$\footnote{https://huggingface.co/mosaicml/mpt-7b}. 
LLaMA \cite{touvron2023llama} models are popular autoregressive text generation models that have obtained outstanding results compared to larger language models. MPT \cite{MosaicMPT7b} is an open-source model developed to get similar performance as LLaMA but available for commercial use. 

\paragraph{Baselines.} We compare \proj against PEFT approaches like Prefix \cite{li-liang-2021-prefix}, Series \cite{pmlr-v97-houlsby19a-s-adapter}, Parallel \cite{pfeiffer-etal-2020-mad-parallel-adapter}, and LoRA \cite{hu2022lora}, using their results reported by LLM Adapters \cite{hu2023llm_adapters}. 
In the case of the GSM8K dataset, we also compare \proj against the results obtained by \citet{kurtic2023sparse}, which uses full fine-tuning.

More details about the experiment implementation are included in Appendix \ref{sec:hyperparameters}.

\subsection{Comparison to LLM-Adapters}

\subsubsection{Math Reasoning}

Table \ref{tab:main_results_llama} shows the comparison of \proj with various adapter approaches.
We fine-tune the sparsified super-adapter network in this experimental scenario utilizing the 10K unified math dataset. Then, the test accuracy on four math reasoning test datasets of the heuristic subnetwork is reported.
As shown in the table, \proj successfully generates subnetworks with higher sparsity levels while demonstrating improvements or marginal drops in accuracy.
At a sparsity level of 40\% \footnote{The actual sparsity is marginally lower than the value in the table (approximately less than 0.5\%), attributed to the introduction of additional parameters for the adapter.} for LLaMA$_{\text{7B}}$, \proj shows performance comparable to PEFT approaches without sparsity. 
Meanwhile, for LLaMA$_{\text{13B}}$, a higher sparsity level of 50\% can be attained while maintaining satisfactory performance.
It is noteworthy that with a sparsity of 40\%, \proj outperforms all PEFT approaches, even surpassing their performance in the absence of any sparsity.

\subsubsection{Commonsense Reasoning}

To further understand \proj' generalizability to other tasks, we fine-tune LLaMA$_{\text{7B}}$ using the unified commonsense dataset from LLM-Adapters \cite{hu2023llm_adapters} using subsets of 15k and 170k samples and evaluate \proj' models at different levels of sparsity on commonsense reasoning datasets. As shown in Table \ref{tab:commonsense170k_results}, at 40\% sparsity, \proj obtains models that outperform the baselines, and at 50\% sparsity, it obtains competitive models, demonstrating the benefits and generalizability of the proposed approach.

\subsection{Comparison to Full Fine-Tuning: MPT with GSM8K}
\label{sec:comparison_to_full_fine_tuning}

\begin{figure}
  \centering
  \includegraphics[width=\linewidth]{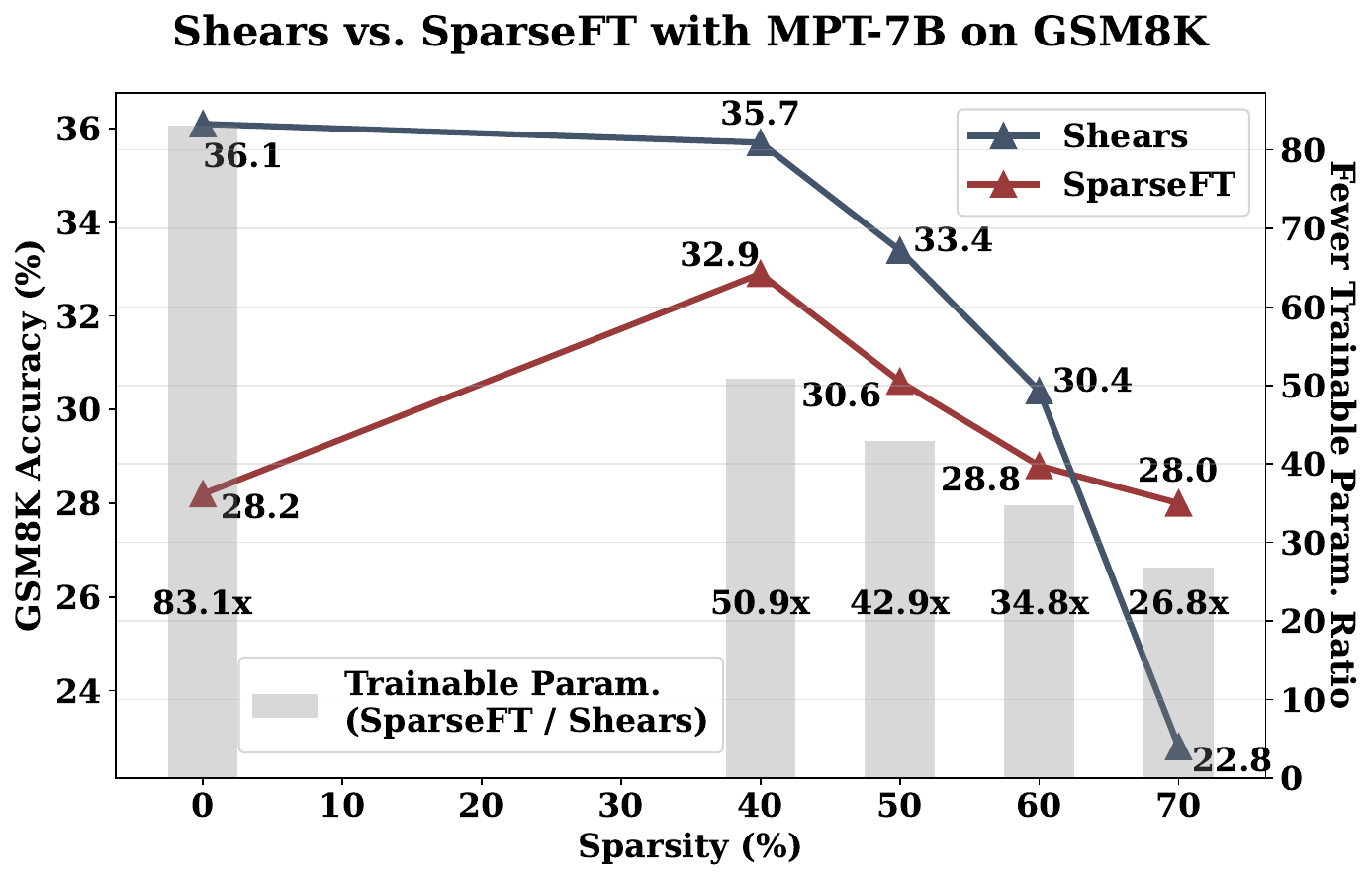}
  \caption{Comparison of \proj and \emph{Sparse Fine-tuning} (SparseFT) \cite{kurtic2023sparse} on the GSM8K test dataset.
  }
  \label{fig:mpt}
\end{figure}

\begin{table}[h] 
\setlength{\belowcaptionskip}{5pt}
\setlength{\tabcolsep}{2.0pt}
\caption{
Comparison of non-zero parameters. 
Acc. represents the average accuracy across all math test datasets.
}
\footnotesize
\centering
\renewcommand\arraystretch{1.2}
\begin{tabular}{llccc}
\toprule
\multirow{2}{*}{\textbf{LLM}} & \multirow{2}{*}{\textbf{Method}}  & \multirow{2}{*}{\textbf{Sparsity}} & \multirow{2}{*}{\textbf{Accuracy(\%)}}  & \textbf{Non-zero} 
\\ 
& &  &  & \textbf{Param.} 
\\ 
\midrule
\multirow{2}{*}{LLaMA$_{\text{7B}}$} & LoRA & - & 46.9 & 6.7B \\
 & \textbf{\proj}    & 50\%  & 45.3 & \textbf{3.5B} \\
\midrule
\multirow{2}{*}{LLaMA$_{\text{13B}}$} & LoRA & - & 51.1 & 13.0B \\
& \textbf{\proj}  & 50\%  & 50.9 & \textbf{6.7B}\\

\bottomrule
\end{tabular}
\label{tab:benefit}
\end{table}

\begin{table*}[hbt]
\setlength{\belowcaptionskip}{5pt}
\setlength{\tabcolsep}{7.5pt}
\caption{Ablation studies for {LLaMA$_{\text{7B}}$}.
For a fair comparison, all ablation experiments with LoRA and NLS tuning applied the same adapter target modules (\textbf{Q},\textbf{ K}, \textbf{V}, \textbf{Up}, and \textbf{Down}).
}

\footnotesize
\centering
\renewcommand\arraystretch{1.2}
\begin{tabular}{lcccccc}
\toprule
\multirow{2}{*}{\textbf{Method}} & \multirow{2}{*}{\textbf{Sparsity}}  & \multicolumn{4}{c}{\textbf{Datasets | Accuracy(\%)}} & \multirow{2}{*}{\textbf{Average}} \\ 
&  & \textbf{GSM8K} & \textbf{AQuA} & \textbf{MAWPS} & \textbf{SVAMP} & \\
\hline\hline
\textit{LLaMA$_{\text{7B}}$:} \\
w/o tune & -  & 11.0 & 24.8 & 3.4 & 2.9 & 10.5 \\
w/ LoRA tune & - & \textbf{37.5} & \textbf{18.9} & 79.0 & \textbf{52.1} & 46.9 \\
w/ NLS tune (\textbf{\proj} w/o sparsity)  & -  & 37.3 & 18.5 & \textbf{82.8} & 49.4 & \textbf{47.0} \\ 

\hline\hline
\multicolumn{7}{l}{\textit{Pruned LLaMA$_{\text{7B}}$:}} \\
w/o tune & 50\%  & 2.5 & 8.7 & 13.0 & 6.5 & 7.7 \\
w/ LoRA tune & 50\%  & 33.8 & 18.1 & \textbf{79.0} & 42.3 & 43.3 \\
w/ NLS tune (\textbf{\proj})  & 50\%  & \textbf{36.1} & \textbf{22.0} & 78.6 & \textbf{44.5} & \textbf{45.3}  \\ 
		
\bottomrule
\end{tabular}
\label{tab:ablation_studies_llama}
\end{table*}

In addition to comparing with other PEFT methods, we conducted experiments to compare \proj and full fine-tuning.
We conduct experiments on a single math reasoning dataset, the GSM8K dataset \cite{cobbe2021training_GSM8K},  generating the MPT$_{\text{7B}}$ super-adapter network. GSM8K can be challenging to LLMs that have not been fine-tuned for this particular task.    
Figure  \ref{fig:mpt} shows a comparison of \proj and recent work by \citet{kurtic2023sparse}, \emph{Sparse Fine-Tuning} (SparseFT). This work employs SparseGPT \cite{frantar-sparsegpt} and full fine-tuning using a novel knowledge distillation strategy. In the case of \proj, we adopt a more efficient approach leveraging unstructured sparsity and only fine-tuning the elastic adapters, which means that \proj incorporates fewer trainable parameters, reducing training and memory costs. SparseFT uses FP32 precision for tuning the whole model weights and employs a knowledge distillation strategy with a more knowledgeable teacher. At the same time, \proj utilizes FP16 precision for pre-trained weights, and the training process does not involve knowledge distillation. 
As shown in the figure, our approach, \proj, outperforms SparseFT across sparsity levels from 0\% to 60\%, which indicates that \proj produces models with similar sparsity but higher accuracy. However, at a sparsity level of 70\%, SparseFT yields higher accuracy but involves the high cost of fine-tuning all the weights in the original model.

\begin{table}[hbt]
\setlength{\belowcaptionskip}{5pt}
\setlength{\tabcolsep}{1pt}
\caption{Ablation studies for {MPT$_{\text{7B}}$}.
Experiments with LoRA and NLS tuning applied the same adapter target modules (\textbf{Q}, \textbf{K}, \textbf{V}, \textbf{O}, \textbf{Up}, and \textbf{Down}). \proj outperforms LoRA with and without the sparsification step.
}

\footnotesize
\centering
\renewcommand\arraystretch{1.2}
\begin{tabular}{lcc}
\toprule
\textbf{Method} & \textbf{Sparsity} & \textbf{Test Accuracy} \\ 
\hline\hline
\textit{MPT$_{\text{7B}}$:} \\
w/o tune & -  &  2.7  \\
w/ LoRA tune & -  & 35.5 \\
w/ NLS tune (\textbf{\proj} w/o sparsity) & -  & \textbf{36.1} \\ 

\hline\hline
\multicolumn{3}{l}{\textit{Pruned MPT$_{\text{7B}}$:}} \\
w/o tune & 40\%  & 2.9 \\
w/ LoRA tune & 40\%  & 33.0 \\
w/ NLS tune (\textbf{\proj})  & 40\%  & \textbf{35.7}  \\  
\cdashline{1-3}
w/o tune & 50\%  & 2.4 \\
w/ LoRA tune & 50\%  & 31.8 \\
w/ NLS tune (\textbf{\proj}) & 50\%  & \textbf{33.4}  \\

\bottomrule
\end{tabular}
\label{tab:ablation_studies_mpt}
\end{table}

\subsection{
Benefits of Sparse Models 
}

Table \ref{tab:benefit} shows the benefits of the high-performing models within the \proj super-adapter network. 
\proj obtains a model with 50\% sparsity that contains $1.91\times$ fewer non-zero parameters with minor drops in accuracy. 
Notably, the model from \proj maintains the adapters unmerged, while the vanilla LoRA adapters are merged with the original model. 
Since the bulk of the model sparsity is concentrated in the frozen weights, combining the adapters will reduce the sparsity levels. Furthermore, benefiting from sparsity, \proj still exhibits notable inference acceleration while maintaining accuracy or experiences only a marginal decrease compared to the vanilla LoRA.

\subsection{Ablation Studies}

Tables \ref{tab:ablation_studies_llama} and \ref{tab:ablation_studies_mpt} illustrate the test accuracy comparison for ablation studies conducted on diverse methods, considering sparsity and various LLMs. The findings indicate that LLaMA$_{\text{7B}}$ and MPT$_{\text{7B}}$ can only effectively handle the challenging downstream datasets with fine-tuning, emphasizing the pivotal role of fine-tuning in these tasks. In the supervised fine-tuning setup, \proj demonstrates some benefits, whether applied to models with or without sparsity.
Specifically, LoRA and \proj perform similarly in the experimental group without sparsity.
However, with 50\% sparsity, \proj outperforms LoRA significantly, highlighting its efficacy in enhancing model performance under sparsity conditions.
This observation underscores that for sparsified models, employing \proj allows for a more substantial maximization of model performance in the supervised fine-tuning setup.

\subsection{Sub-Adapter Configuration Search}

Table \ref{tab:sub_adapters} demonstrates the accuracy range of the search space of sub-adapter configurations. Since the sparsified weights of the model remain frozen, the search for the configuration of the attached adapter in \proj is significantly smaller than the search space in general neural architecture search.  
Studies indicate a narrow accuracy range, with the difference in accuracy between the minimal and the maximal sub-adapter configuration being only a single accuracy percentage point.
The heuristic obtained in O(1) already gives us a reliable indication of the quality of the sub-adapters around the mid-configuration space. If the user has the budget, a more refined sub-adapter configuration can be searched using a cost-effective hill-climbing strategy that is cheaper than other methods, e.g., evolutionary search with RNSGA-II.

\begin{table}[h]
\setlength{\belowcaptionskip}{5pt}
\setlength{\tabcolsep}{4pt}
\caption{
Comparison of various sub-adapter networks and the method used to obtain them from the LLaMA$_{\text{7B}}$ + \proj super-adapter network.
Accuracy represents the average accuracy across all math test datasets.
}
\footnotesize
\centering
\renewcommand\arraystretch{1.2}
\begin{tabular}{lclc}
\toprule
\textbf{Method} & \textbf{Sparsity} &\textbf{Sub-Adapter} & \textbf{Accuracy (\%)} \\ 
\midrule
LoRA & - & - & 46.9 \\
\midrule
\multirow{5}{*}{\textbf{\proj}} & \multirow{5}{*}{50\%} &  Maximal  & 44.5 \\
&  &  Heuristic  & 45.3 \\
&  &  Hill-climbing  & \textbf{45.9} \\
&  &  RNSGA-II  & 45.7 \\
&  &  Minimal  & 43.5 \\

\bottomrule
\end{tabular}
\label{tab:sub_adapters}
\end{table}

%% file: content/5_conclusion.tex
\section{Conclusion}
\label{sec:conclusion}

This paper presents \textbf{\proj}, a practical and novel solution for real-world applications to sparsifying weight-sharing super-networks of elastic adapters (super-adapters). By incorporating elastic LoRA adapters into the sparsified base model, \proj can fine-tune LLMs without sacrificing the sparsity obtained from the original model weights and produces sparse models with improvements or minor drops in accuracy and a fraction of the cost compared to other approaches. The increase in sparsity 
can result in significant speedup when using runtimes that take advantage of these patterns. Ablation studies show that combining 
sparsified models with elastic low-rank adapters 
yields better results than using LoRA adapters alone. Models and code are available at \href{https://github.com/IntelLabs/Hardware-Aware-Automated-Machine-Learning}{https://github.com/IntelLabs/Hardware-Aware-Automated-Machine-Learning}.

\section*{Ethical Considerations and Limitations}

The significant size of recent large language models has brought challenges for fine-tuning and deployment.  Users with proprietary data must spend considerable time and resources adjusting LLMs' weights to improve their performance on custom tasks. In a world with limited resources, it is an ethical concern to find approaches that reduce the requirements of training and fine-tuning LLMs. Although \proj significantly reduces this process's requirements, more work is needed to address this issue.  There is also the need for more research on the inherent limitations of LLMs. Their results and decisions should be carefully audited when they can affect customers' lives, who might need to be made aware of the depths and gaps in understanding that LLM researchers still have. Our goal is to make these models more efficient. However, efficiency is not the end of the story, and the above limitations should be considered when using or sharing LLMs.

\section*{Acknowledgments}

We are grateful to Michael Beale from Intel Labs, who helped us set up the infrastructure for sharing our models during the review stage and the final release and guided us through open-sourcing our compressed models. We also thank the anonymous reviewers for their insightful suggestions, which helped us improve the paper.

%% file: content/appendix.tex
\input{content/appendix_A_related_work}

\section{Hyperparameters}
\label{sec:hyperparameters}

The hyperparameters of our approach under different LLMs are listed in Table \ref{tab:hyperparameters_llama_math}, Table \ref{tab:hyperparameters_llama_commonsense}, and Table \ref{tab:hyperparameters_mpt}.

\begin{table*}[hbt]
\setlength{\belowcaptionskip}{5pt}
\setlength{\tabcolsep}{7.5pt}
\caption{Hyperparameters for LLaMA-series models with the math reasoning dataset.}

\footnotesize
\centering
\renewcommand\arraystretch{1.2}
\begin{tabular}{lccccccc}
\toprule
Model & LLaMA$_{\text{7B}}$ & LLaMA$_{\text{7B}}$  & LLaMA$_{\text{13B}}$ & LLaMA$_{\text{13B}}$ \\ 
Sparsity & 40\% & 50\% & 40\% & 50\% \\
\cdashline{1-5}
Epoch & 4 & 3 & 3 & 3 \\
Batch size & 16 & 16 & 16 & 16 \\
Learning rate & 3e-4 & 3e-4 & 3e-4 & 3e-4 \\
LoRA alpha & 64 & 64 & 64 & 64 \\
LoRA target modules  & Q, K, V, Up, Gate, Down & Q, K, V, Up, Down & Q, K, V, Up, Down & Q, K, V, Up, Down \\
Low-rank Search Space & [32, 24, 16] & [32, 24, 16] & [32, 24, 16] & [32, 24, 16] \\
\bottomrule
\end{tabular}
\label{tab:hyperparameters_llama_math}
\end{table*}

\begin{table*}[hbt]
\setlength{\belowcaptionskip}{5pt}
\setlength{\tabcolsep}{5.5pt}
\caption{Hyperparameters for LLaMA$_{\text{7B}}$ with the commonsense reasoning dataset.}

\footnotesize
\centering
\renewcommand\arraystretch{1.2}
\begin{tabular}{lccccccc}
\toprule
Train set size & 15k & 15k & 170k & 170k \\ 
Sparsity & 40\% & 50\% & 40\% & 50\% \\
\cdashline{1-5}
Epoch & 3 & 3 & 3 & 5 \\
Batch size & 16 & 16 & 16 & 16 \\
Learning rate & 3e-4 & 3e-4 & 3e-4 & 3e-4 \\ 
LoRA alpha & 64 & 64 & 64 & 64 \\
LoRA target modules  & Q, K, V, Up, Down & Q, K, V, Up, Gate, Down & Q, K, V, Up, Gate, Down & Q, K, V, Up, Down \\
Low-rank Search Space & [32, 24, 16] & [32, 24, 16] & [32, 24, 16] & [32, 24, 16] \\
\bottomrule
\end{tabular}
\label{tab:hyperparameters_llama_commonsense}
\end{table*}

\begin{table*}[hbt]
\setlength{\belowcaptionskip}{5pt}
\setlength{\tabcolsep}{5.5pt}
\caption{Hyperparameters for MPT$_{\text{7B}}$ with GSM8K.}

\footnotesize
\centering
\renewcommand\arraystretch{1.2}
\begin{tabular}{lccccccc}
\toprule
Sparsity & 40\% & 50\% & 60\% & 70\%\\
\cdashline{1-6}

Epoch & 4 & 5 & 5 & 8  \\
Batch size & 16 & 16 & 16 & 16\\
Learning rate & 5e-4 & 3e-4 & 3e-4 & 3e-4 \\
LoRA alpha & 64 & 64 & 64 & 64 \\
LoRA target modules  & Q, K, V, O, Up, Down & Q, K, V, O, Up, Down & Q, K, V, O, Up, Down & Q, K, V, O, Up, Down \\
Low-rank Search Space & [32, 24, 16] & [32, 24, 16] & [32, 24, 16] & [32, 24, 16]\\
\bottomrule
\end{tabular}
\label{tab:hyperparameters_mpt}
\end{table*}

%% file: content/appendix_A_related_work.tex
\section{Related Work}
\label{sec:related_work}

\paragraph{Neural Architecture Search (NAS)} Given a set of possible deep learning architecture configurations, a NAS algorithm discovers high-performing configurations. They often update the model weights, yielding a trained model ready for deployment. Research in NAS has increased dramatically in the past few years \cite{nas1000}, making these techniques highly popular with practitioners engaged in model optimization and compression. One-shot weight-sharing neural architecture search has been demonstrated to be a practical class of NAS algorithms with savings in memory and additional storage since they construct a super-network that contains a large number of subnetworks \cite{cai2020once, Bignas}. In this case, the objective of the NAS algorithm is to train and identify an outstanding subnetwork, frequently representing a compressed version of the original model. 
Our approach, \proj, differs from traditional NAS in that we do not attempt to find a better, more efficient neural architecture using the original model as a reference. \proj freezes the original model and attaches elastic low-rank adapters, directing the NAS mechanisms only to these adapters, termed neural low-rank adapter search (NLS).

\paragraph{Elastic Adapters} 
PEFT \cite{ding2022delta} has become a popular method for fine-tuning large models. Recently, there has been work on making the adapters in PEFT elastic, aiming to find the optimal adapter configuration through a search process.
AutoPEFT \cite{zhou2023autopeft} automatically applies elastic serial adapters, parallel adapters, and prefix-tuning into the small language model like BERT to identify the optimal adapter class and its configuration within these elastic modules.
LoNAS \cite{lonas2024} introduces elasticity to the low-rank adapters and pre-trained weights in LLM, enabling them to adopt various configurations. 
This feature effectively generates a search space conducive to exploring using weight-sharing neural architecture search (NAS). 
In our approach, \proj only makes the LoRA adapters of the sparsified model elastic, ingeniously combining both model sparsification and elastic adapters to elicit optimal performance in the sparsified model.

\paragraph{Sparsity and Pruning} Pruning the weights of a neural network is a popular technique for model compression. The most common approach of element-wise pruning uses the \emph{magnitude} of the weights and a thresholding function that zeroes out the weights below a threshold. Weight \emph{magnitude pruning} is ineffective when applied to LLMs \cite{frantar-sparsegpt}. One possible reason is the existence of outlier features when models reach several billion parameters \cite{dettmers2022gptint}. Alternative approaches have been proposed to measure the importance of the weights. For instance, first-order approaches use several iterations to update the weights, e.g., Movement Pruning \cite{sanh2020movement} and  SparseGPT \cite{frantar-sparsegpt}. These approaches have also improved weight-sharing NAS \cite{eftnas2024}. Unfortunately, using weight updates for LLM pruning requires a significant computational cost. Recently, efficient approaches have been proposed to achieve high degrees of sparsity with a single forward pass of $N$ samples. For example, Wanda \cite{sun2023wanda} is a simple but effective sparsification method that determines which parameters to zero out by the importance of weights based on both the weights and the activations.
LLM-Pruner \cite{ma2023llmpruner} is proposed to compress LLMs in a task-agnostic manner \cite{ma2023llmpruner}. This approach produces good zero-shot results after applying structured pruning on the targe LLM. 
Unlike these approaches, \proj is designed for specific task fine-tuning scenarios, which can obtain higher levels of unstructured sparsity while improving or with minor drops in accuracy by combining unstructured sparsity with neural low-rank adapter search (NLS). 

\paragraph{Sparsity and Fine-Tuning} SparseFT \cite{kurtic2023sparse} uses SparseGPT \cite{frantar-sparsegpt} to sparsify the model and then fine-tunes all the weights of the model using a novel knowledge distillation technique (see section \ref{sec:comparison_to_full_fine_tuning}). Unlike SparseFT, \proj does not use knowledge distillation and fine-tunes only a tiny set of weights in elastic low-rank adapters. 
Our approach necessitates updating only a fraction of the total parameters, thereby reducing memory and computing demands during training while enhancing accuracy.